\documentclass[11pt,letterpaper]{article}
\usepackage{emnlp2016}
\usepackage{times}
\usepackage{latexsym}

\usepackage{todonotes}
\usepackage{bm}
\usepackage{amsmath}
\usepackage{graphicx}
\usepackage{float}
\usepackage[font=footnotesize,labelfont=bf]{caption}
\usepackage{subcaption}
\usepackage{booktabs}
\usepackage{amssymb}

\usepackage{mathtools}
\DeclarePairedDelimiter{\norm}{\lVert}{\rVert}

\graphicspath{{./figures/}}
\newcommand{\word}[1]{\emph{#1}}
\newcommand{\tech}[1]{\emph{#1}}
\newcommand{\set}[1]{\left\{#1\right\}}

\usepackage{color} 
\usepackage{url} 

\DeclareMathOperator{\softmax}{softmax}
\newcommand{\open}[1]{\left(#1\right)} 

\makeatletter

\def\citealt{\def\citename##1{{\frenchspacing##1} }\@internalcitec}

\def\@citexc[#1]#2{\if@filesw\immediate\write\@auxout{\string\citation{#2}}\fi
  \def\@citea{}\@citealt{\@for\@citeb:=#2\do
    {\@citea\def\@citea{;\penalty\@m\ }\@ifundefined
       {b@\@citeb}{{\bf ?}\@warning
       {Citation `\@citeb' on page \thepage \space undefined}}%
{\csname b@\@citeb\endcsname}}}{#1}}

\def\@internalcitec{\@ifnextchar [{\@tempswatrue\@citexc}{\@tempswafalse\@citexc[]}}

\def\@citealt#1#2{{#1\if@tempswa, #2\fi}}

\makeatother

\emnlpfinalcopy

\def\emnlppaperid{***}

\title{On the Effective Use of Pretraining for Natural Language Inference}

\author{Ignacio Cases\\
  Stanford Linguistics\\
  {\tt cases@stanford.edu}
  \And
  Minh-Thang Luong\\
  Stanford Computer Science\\
  {\tt lmthang@stanford.edu}
  \And
  Christopher Potts\\
  Stanford Linguistics\\
  {\tt cgpotts@stanford.edu}}

\date{Spring 2016}

\begin{document}

\maketitle


\begin{abstract}
  Neural networks have excelled at many NLP tasks, but there remain open
questions about the performance of pretrained distributed word
representations and their interaction with weight initialization and
other hyperparameters. We address these questions empirically using
attention-based sequence-to-sequence models for natural language
inference (NLI). Specifically, we compare three types of embeddings:
\tech{random}, \tech{pretrained} (GloVe, word2vec), and
\tech{retrofitted} (pretrained plus WordNet information). We show that
pretrained embeddings outperform both random and retrofitted ones in a
large NLI corpus. Further experiments on more controlled data sets
shed light on the contexts for which retrofitted embeddings can be
useful. We also explore two principled approaches to initializing the
rest of the model parameters, Gaussian and orthogonal, showing that
the latter yields gains of up to 2.9\% in the NLI task.

\end{abstract}

\section{Introduction}
Unsupervised pretraining of supervised neural network inputs has
proven valuable in a wide range of tasks, and there are strong
theoretical reasons for expecting it to be useful, especially where
the network architecture is complex and the available training data is
limited \cite{erhan2009difficulty,erhan2010does}. However, for
natural language processing tasks, the results for pretraining remain
mixed, with random input initializations sometimes even appearing
superior. This is perhaps an unexpected result given the comparatively
small size of most labeled NLP datasets and the consistent structure
of natural language data across different usage contexts.

In this work, we trace the variable performance of pretraining in NLP
to the ways in which these dense, highly structured representations
interact with other properties of the network, especially the learning
rate parameter and the weight initialization scheme. The basis for
these experiments is the Stanford Natural Language Inference (SNLI)
corpus \cite{bowman2015}, which contains about 570k sentence pairs
labeled for entailment, contradiction, and semantic independence. SNLI
is an ideal choice for a number of reasons: it is one of the larger
human-annotated resources available today, it is oriented towards a
task that has wide applicability in NLP \cite{Dagan06}, and its
relational structure is similar to what one finds in datasets for
machine translation, paraphrase, and a number of other tasks.

\begin{figure}[tp]
  \centering
  \includegraphics[width=0.45\textwidth, clip=true, trim= 0 0 0 0]{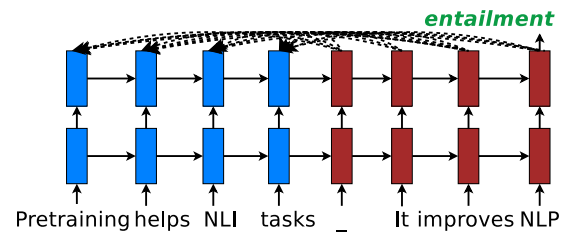}
  \caption{An example of an attention-based sequence-to-sequence model 
    that classifies the relationship between a \emph{premise}, ``pretraining helps NLI'', 
    and a \emph{hypothesis}, ``it improves NLP'', as  \emph{entailment}. 
    Dotted arrows illustrate the role of attention
    in selecting relevant information from the premise.}
  \label{f:nli}
\end{figure}

Our approach utilizes attention-based sequence-to-sequence models as
illustrated in Figure~\ref{f:nli}. Using SNLI, we test the performance
of three embedding classes: \tech{random}; \tech{pretrained} -- GloVe;
\cite{pennington2014} and \tech{word2vec} \cite{Mikolov-etal:2013:ICLR} -- as well as
\tech{retrofitted}, i.e., pretrained embeddings plus WordNet
information \cite{faruqui2014}. Via large hyperparameter searches, we
find that pretrained GloVe and word2vec significantly outperform random as long
as the network is properly configured. As part of these experiments,
we systematically evaluate two approaches to weight initialization:
\tech{Gaussian} \cite{glorot2010} and \tech{orthogonal} \cite{saxe2014}, finding that
the latter yields consistent and substantial gains.

Surprisingly, retrofitting degrades performance, even though the
WordNet information that this brings in should be well-aligned with
the SNLI labels, given the close association between the WordNet
nominal hierarchy and the nature of the NLI task. In a second set of
experiments, we show that retrofitting \emph{can} be
helpful on word-level NLI-like tasks. Indeed, in these experiments, its
performance is far superior to the other options.  This suggests a
hypothesis: if the inputs carry extensive information that word $A$
entails word $B$, might that make it difficult to adjust to the fact
that, for example, $\word{not}~B$ therefore entails $\word{not}~A$?
To test this hypothesis, we rely on a rich theory of negation derived
from the natural logic of \newcite{maccartney2009a} to incrementally
add compositional complexity to the work-level task. We find that, as
semantic complexity goes up, the performance of retrofitting declines.

Overall, these results indicate that working with pretrained inputs
requires care but reliably improves performance of the NLI task.

\section{Motivation}\label{sec:motivation}
%
%
%

Unsupervised pretraining has been central to recent progress in deep
learning \cite{lecun2015}. \newcite{hinton2006} and
\newcite{bengio2007} show that greedy layer-wise pretraining of the
inner layers of a deep network allows for a sensible initialization of
the parameters of the network. Similar strategies were central to some
recent major success stories of deep networks (e.g.,
\citealt{mohamed2012,dahl2012,sermanet2013}). \newcite{dai2015semi}
show that initializing a recurrent neural network (RNN) with an
internal state obtained in a previous phase of unsupervised
pretraining consistently results in better generalization and
increased stability. More generally, the dense representations
obtained by methods like GloVe and word2vec have, among other things,
enabled the successful use of deep, complex models that could not
previously have been optimized effectively.

Some recent work calls into question the importance of pretraining,
however. One can train deep architectures without pretraining and
achieve excellent results \cite{martens2010,sutskever2013}, by using
carefully chosen random initializations and advanced optimization
methods, such as second order methods and adaptive gradients. It has
been shown that randomly initialized models can find the same local
minima found by pretraining if the models are run to convergence
\cite{martens2010,chapelle2011,saxe2015}. One might conclude from
these results that pretraining is mostly useful only for preventing
overfitting on small datasets \cite{bengio2013,lecun2015}.

This conclusion would be too hasty, though. New analytical approaches
to studying deep \emph{linear} networks have reopened questions
about the utility of pretraining. Drawing on a rich literature,
\newcite{saxe2014} and \newcite{saxe2015} show that pretraining still
offers optimization and generalization advantages when it is combined
with standard as well as advanced optimization methods. 
There are also clear practical advantages to pretraining: faster
convergence (with both first- and second-order methods) and better
generalization. For very large datasets (close enough to the asymptotic
limit of infinite data), some of training time will actually be spent
in achieving the effects of pretraining. Avoiding such redundant
effort might be crucial. For small or medium-sized datasets, this
might not be an option, in which case pretraining is clearly the right
choice.

For the most part, results achieved using random initial conditions
have not been systematically compared against the use of pretrained
initial conditions, even where pretraining is thought to be
potentially helpful \cite{krizhevsky2012,sutskever2013}. This is
perhaps not surprising in light of the fact that systematic
comparisons are challenging and resource-intensive to make. As
\newcite{saxe2015} note, the inputs interact in complex ways with all
the other network parameters, meaning that systematic comparisons
require extensive hyper-parameter searches on a variety of
training-data sources. The next section defines our experimental
framework for making these comparisons in the context of NLI. The
findings indeed reveal complex dependencies between data, network
parameters, and input structure.

\section{Model architecture}\label{sec:rnn}
Attention-based sequence-to-sequence (seq2seq) models have been effective for
many NLP tasks, including machine translation
\cite{bahdanau2014,luong2015}, speech recognition \cite{chorwoski15},
and summarization \cite{rush2015}.  In NLI, variations of
this architecture have achieved exceptional results
\cite{rocktaschel2015,wang2015,cheng2016}. It is therefore our architecture of choice. 


A seq2seq model generally consists of two components, an
\tech{encoder} and a \tech{decoder}
\cite{cho2014learning,sutskever2014}, each of which is an RNN. In NLI,
each training example is a triple (\emph{premise}, \emph{hypothesis},
\emph{label}). The encoder builds a representation for the premise
and passes the information to the decoder. The decoder reads
through the hypothesis and predicts the assigned label, as illustrated
in Figure~\ref{f:nli}.

Concretely, for both the encoder and the decoder, the hidden state at time $t$  is derived from the previous state
$\bm{h}_{t{-}1}$ and the current input $\bm{x}_t$ as follows: 
\begin{align}
\bm{h}_t = f\open{\bm{h}_{t{-}1}, \bm{x}_t}
\end{align}
Here, $f$ can take different forms, e.g., a vanilla
RNN or a long short-term memory network (LSTM; \citealt{hochreiter1997}). In all our models, we choose to use the
multi-layer LSTM architecture described by \newcite{zaremba14}. At the bottom
LSTM layer, $\bm{x}_t$ is the vector representation for the word at time $t$,
which we look up from an \emph{embedding matrix} (one for each encoder and
decoder). This is where we experiment with different embedding classes (Section~\ref{sec:embedding-classes}).

On the hypothesis side, the final state at the top LSTM layer
$\bm{h}_{H}$ is passed through a softmax layer to compute the probability of
assigning a label $l$:
\begin{equation}
p\open{l|\mbox{premise}, \mbox{hypothesis}} = \softmax\open{\bm{h}_{H}}
\label{e:softmax}
\end{equation}
which we use to compute the cross-entropy loss to minimize during training. At
test time, we extract the most likely label for each sentence pair. 

One key addition to this basic seq2seq approach is the use of \emph{attention mechanisms}, which have proven to be valuable for NLI \cite{rocktaschel2015}. The
idea of attention is to maintain a random access memory of all
previously-computed hidden states on the encoder side (not just the last one).
Such memory can be referred to as the decoder hidden states are built
up, which improves learning, especially for long sentences. In our models,
we follow \newcite{luong2015} in using
the \tech{local-p} attention mechanism and the \tech{dot-product}
attention scoring function.

Apart from the embedding matrices, which can be either pretrained or
randomly initialized, all other model parameters (LSTM, softmax,
attention) are randomly initialized. We examine various important
initialization techniques in Section~\ref{sec:hyperparameters}.

\section{Methods for Effective Use of Pretraining} 
This section describes our approach to comparatively evaluating
different input embedding classes. We argue that, for each 
embedding class, certain preprocessing steps can be essential,
and that certain hyperparameters are especially sensitive to
the structure of the inputs.

\subsection{Embedding Classes}\label{sec:embedding-classes}

Our central hypothesis is that the structure of the inputs is
crucial. To evaluate this idea, we test random, GloVe
\cite{pennington2014}, and word2vec
\cite{Mikolov-etal:2013:ICLR,Goldberg:Levy:2014} inputs, as well
variants of GloVe and word2vec that have been retrofitted with WordNet
information \cite{faruqui2014}.

For the definitions of the GloVe and word2vec models, we refer to the
original papers.\footnote{We used the publicly released embeddings, trained with Common Crawl 840B tokens for GloVe (\url{http://nlp.stanford.edu/projects/glove/}) and Google News 42B for word2vec \url{https://code.google.com/archive/p/word2vec/}. Although the training data sizes are notably different, our goal is not to compare these two models directly. The retrofitting algorithm can be found at \url{github.com/mfaruqui/retrofitting}.} The further step of retrofitting is defined as
follows. A $d$-dimensional vector for a word $j$ drawn from a
vocabulary $V$, $\bm{x}_j$, is retrofitted into a vector
$\bm{\tilde{x}}_j$ by minimizing an objective that seeks to keep the
retrofitted vector $\bm{\tilde{x}}_j$ close to the original, but
changed in a way that incorporates a notion of distance defined by a
source external to the embedding space $E$, such as WordNet
\cite{WordNet98}. The objective of \newcite{faruqui2014} is defined
as follows:
\begin{equation}
\sum_{j \in V} 
  \Bigg\{ 
    \alpha_j \norm{\bm{\tilde{x}}_j - \bm{x}_j}^2 + 
    \sum_{i,j \in E} \beta_{ji} \norm{\bm{\tilde{x}}_j - \bm{\tilde{x}}_i}^2 
  \Bigg\}
\end{equation}
where $\alpha$ and $\beta$ are parameters that control the relative strength of each contribution. When the external source is WordNet, $\alpha_j$ is set to $1$ and $\beta_{ji}$ is the inverse of the degree of node $j$ in WordNet. In our experiments, we used the off-the-shelf algorithm provided by
the authors, keeping the pre-established parameters, with synonyms, hypernyms and hyponyms as connections in the graph \cite{faruqui2014}.

\subsection{Embedding Preprocessing}

GloVe and word2vec embeddings can generally be used directly, with no
preprocessing steps required. However, retrofitting generates vectors
that are not centered and with amplitudes
one magnitude lower than those of GloVe or word2vec. We therefore took
two simple preprocessing steps that helped performance.
First, each dimension was {\it mean centered}. This step makes the
distribution of inputs more similar to the distribution of weights,
which is especially important because of the coupling of inputs and
parameters alluded to at the end of Section~\ref{sec:motivation}.
Second, each dimension was {\it rescaled} to have a standard
deviation of one. Rescaling is usually applied to inputs showing high
variance for features that should contribute equally. 
It is particularly useful when embeddings have either very large or very small amplitudes -- the latter is what we often observe in retrofitted vectors.
The overall effect of these two preprocessing steps is to make the summary
statistics of all our input vectors similar.

\subsection{Hyperparameter Search}\label{sec:hyperparameters}

We expect
the performance of different inputs to be sensitive not only to
the data and objective, but also to the hyperparameter settings. To
try to find the optimal settings of these parameters for each kind of
input, we rely on random search through hyperparameter space.
\newcite{bergstra2012} argue convincingly
that random search is a reliable approach because the hyperparameter
response function has a low effective dimensionality, i.e., it is more
sensitive along some dimensions than others.

Our search proceeds in two stages. First, we perform a coarse search seeking out the dimensions
of greater dependency in the hyperparameter space by training the
model for a small number of epochs (usually just one). Second, we
perform a finer search across these dimensions, holding the other
parameters constant. This finer search is iteratively
\tech{annealed}: on each iteration, the hypercube that delimits the
search is recentered around the point found in the previous iteration
and rescaled with a factor (set to $0.9$ in all our
experiments).

Our hyperparameter searches focus on the learning rate, the
initialization scheme, and the initialization constant. For the
initialization scheme, we choose between \tech{Gaussian
  initialization}, with the corrective multiplicative factor
$1/\sqrt 2^L$, where $L$ is the number of layers
\cite{glorot2010,he2015}, and the \tech{orthogonal random
  initialization} introduced by \newcite{saxe2014}.\footnote{We
  operate on square matrices, e.g., for an LSTM $4d{\times}2d$
  parameter matrix, we orthogonalize 8 $d{\times}d$ sub-matrices.}
The initialization constant is a multiplicative constant $\kappa$
applied to each of the initialization schemes. Other hyperparameters
were set more heuristically. See Section~\ref{sec:SNLI:setting} and
the Appendix~\ref{appendix:hyper} for more details.

\section{Experiments}
\begin{figure*}[tp]
    \centering
    \begin{subfigure}[b]{0.3\textwidth}
        \centering
        \includegraphics[width=\textwidth]{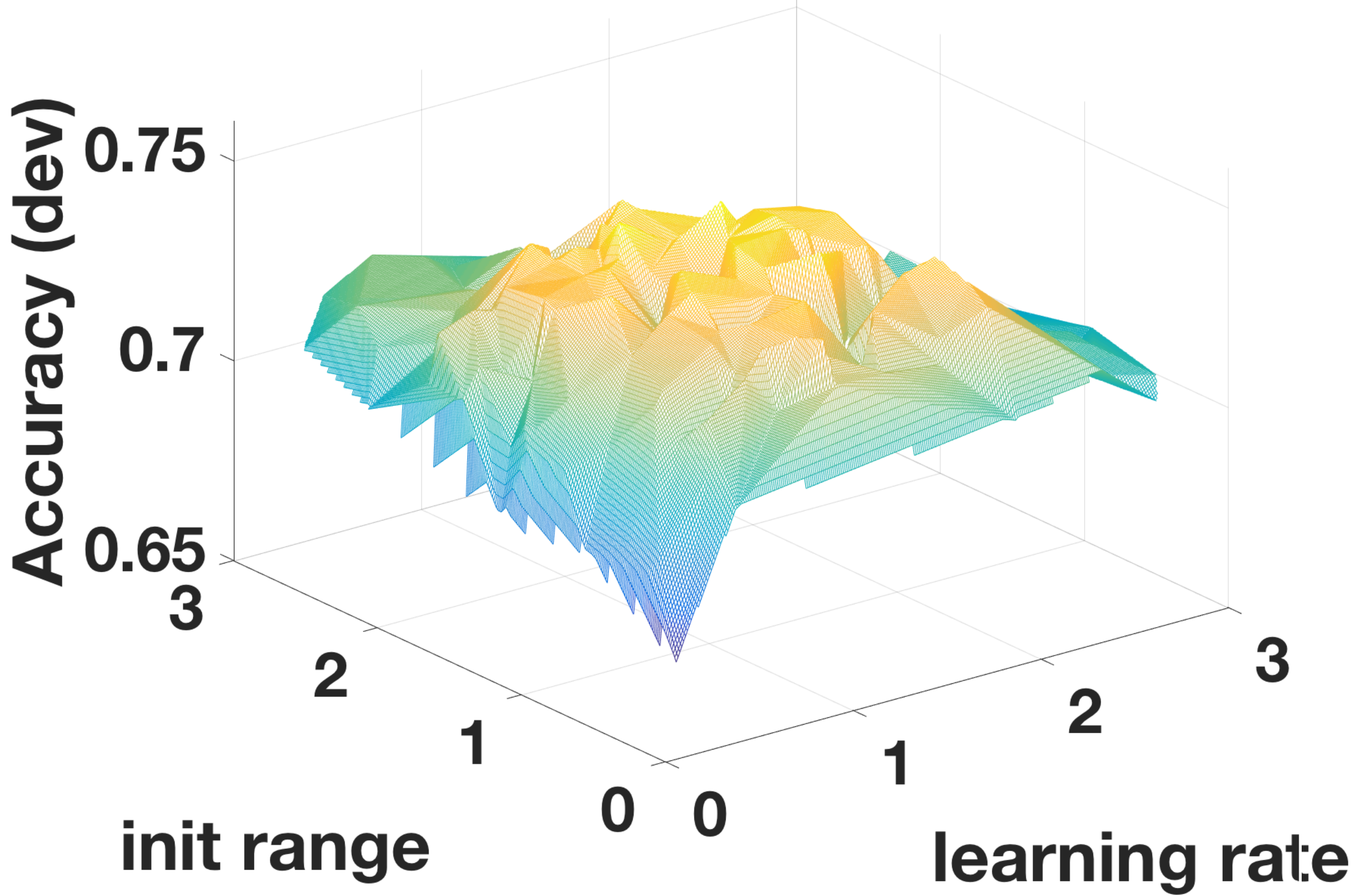}
        \caption{GloVe}
        \label{fig:five over x}
    \end{subfigure}
    \hfill
    \begin{subfigure}[b]{0.3\textwidth}
        \centering
        \includegraphics[width=\textwidth]{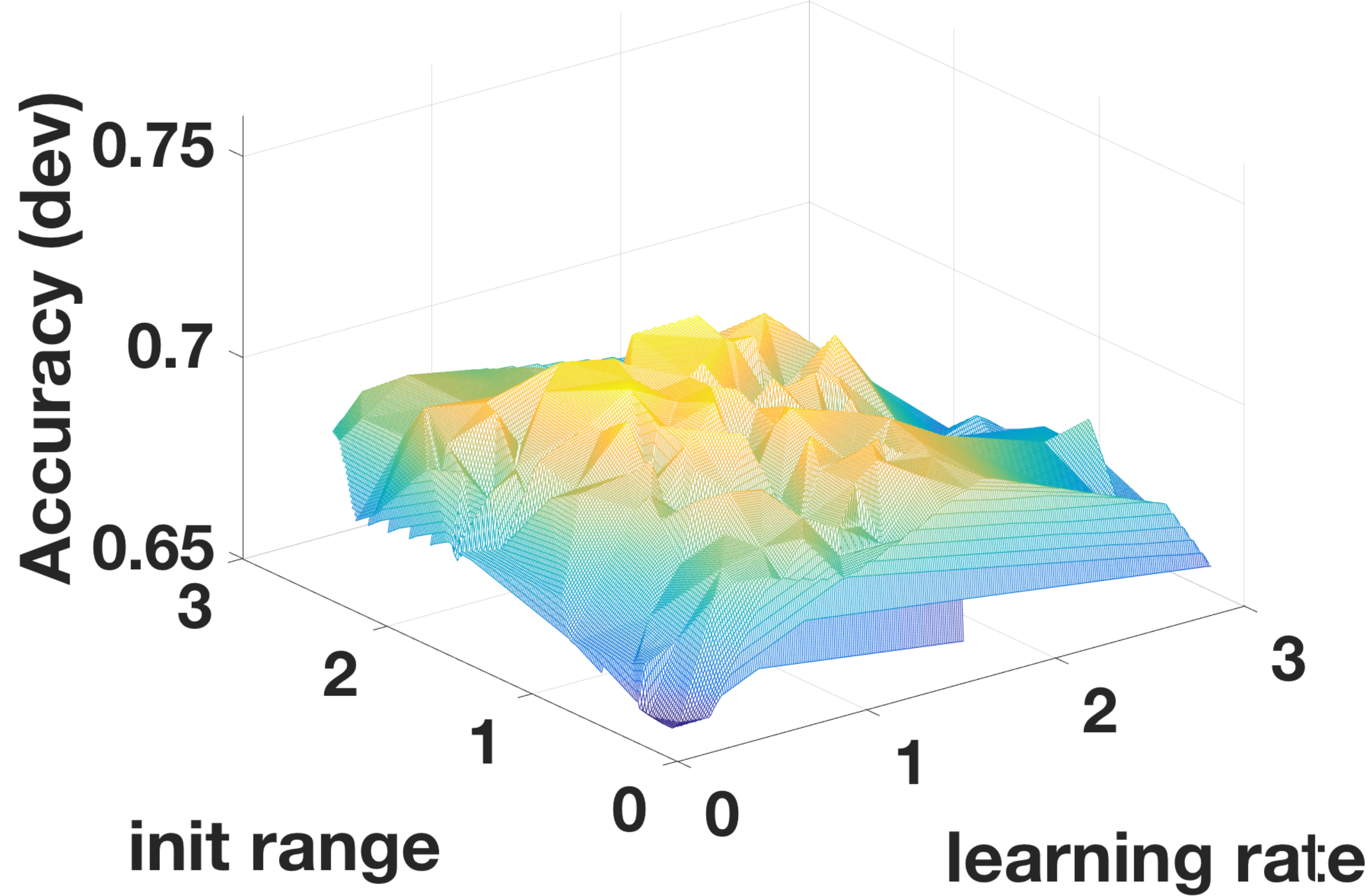}
        \caption{retrofitted GloVe}
        \label{fig:three sin x}
    \end{subfigure}
    \hfill
    \begin{subfigure}[b]{0.3\textwidth}
        \centering
        \includegraphics[width=\textwidth]{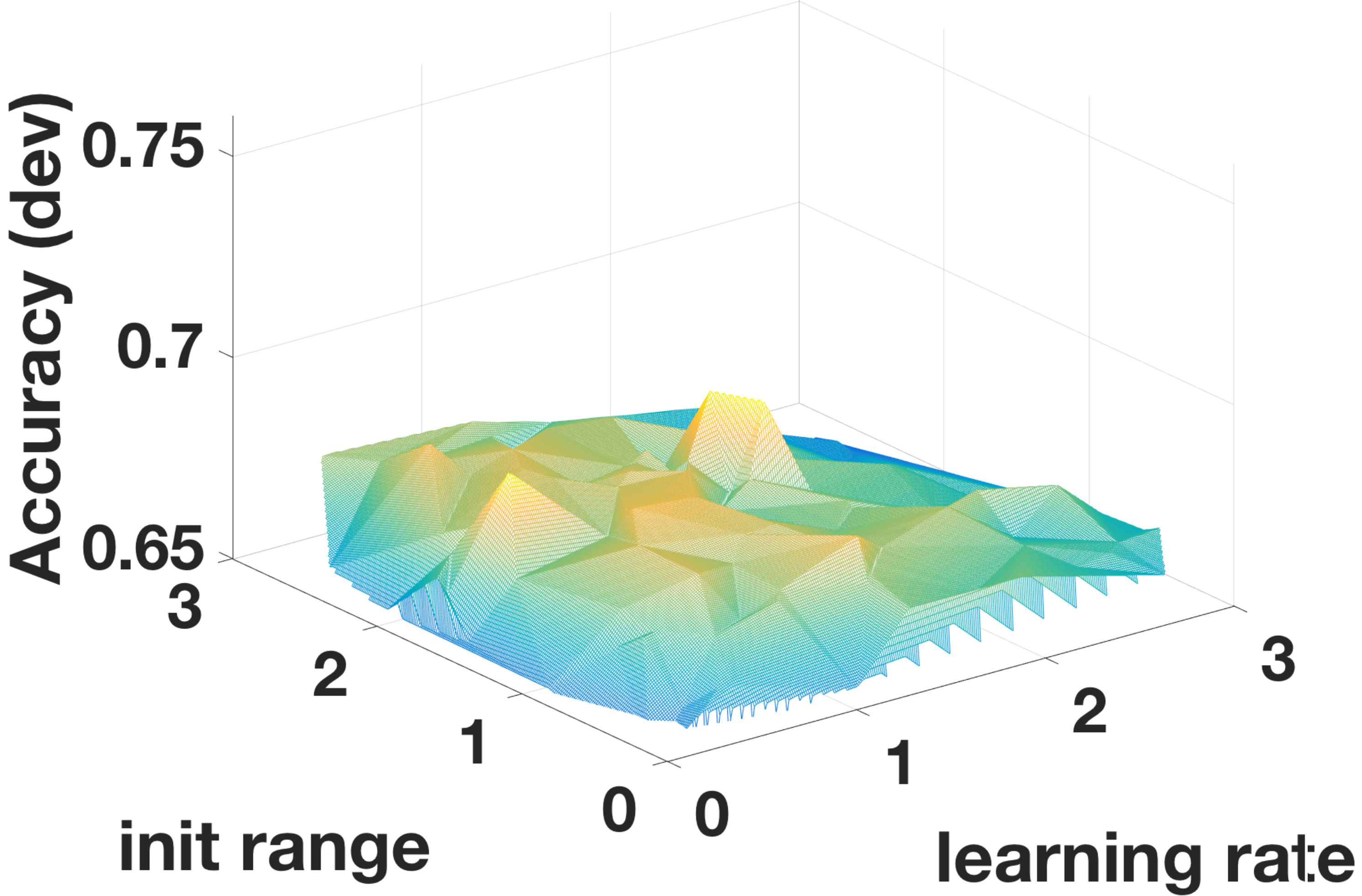}
        \caption{random}
        \label{fig:y equals x}
    \end{subfigure}
    \caption{Hyperparameter landscape for Gaussian weight initialization in 
      the subspace spanned by our initialization and learning parameter ranges, 
      for three 300d embeddings. 
      Word2vec and word2vec retrofitted resulted in similar landscapes as 
      GloVe and GloVe retrofitted, respectively, 
      and are not shown here for reasons of space.
      The figures were generated using a mesh interpolation over the random search points.}
    \label{fig:hyper-search-landscapes}
\end{figure*}

Our primary experiments are with the Stanford NLI (SNLI) corpus
\cite{bowman2015}. This is one of the largest purely human-annotated
datasets for NLP, which means that it offers the best chance for
random initializations to suffice (Section~\ref{sec:motivation}). In
addition, to gain analytic insights into how the inputs
behave, we report on a series of evaluations using
a smaller word-level dataset extracted from WordNet \cite{bowman2014}
and a controlled extension of the dataset to include negation. These
experiments help us further understand the variable performance of
different pretraining methods.

Ideally, we would evaluate our full set of input designs on a wide
range of different corpora from a variety of tasks. Unfortunately, the
hyperparameter searches that are crucial to our empirical argument are
extremely computationally intensive, ruling out comparable runs on
multiple large datasets. Given these constraints, the NLI task seems
like an ideal choice. As noted in Section~\ref{sec:rnn}, it shares
characteristics with paraphrase detection and machine translations, so we hope
the results will transfer. More
importantly, it is simply a challenging task because of its dependence
on lexical semantics, compositional semantics, and world knowledge
\cite{maccartney2009b,dagan2010}.  

\subsection{Stanford Natural Language Inference}
In our SNLI experiments, we use the deep LSTM-based recurrent neural
network with attention introduced in Section~\ref{sec:rnn}.
The training set for SNLI contains $\approx$550k sentence
pairs, and the development and test sets each contain $\approx$10k
pairs, with a joint vocabulary count of 37,082 types. We use only the training and development sets here; given our
goal of comparing the performance of different methods, we need not use
the test set.

\label{sec:SNLI:setting}

As described above, we evaluated five families of initial embeddings:
random, GloVe, word2vec, retrofitted GloVe, and retrofitted word2vec.
For each, we set the learning rate, initialization constant $\kappa$, and initialization scheme
(Gaussian or orthogonal) using the search procedure described in
Section~\ref{sec:hyperparameters}.\footnote{Both the learning rate and
initialization constant are chosen log-uniformly from
$[0.001, 3]$} Other hyperparameters found to be
less variable are listed in the Appendix~\ref{appendix:hyper}; their
values were fixed based on the results of a coarse search.

Although the hyperparameter landscape provides useful insights about
the regions where the models perform adequately, an appropriate
assessment of the dynamics of the models and their initializations
requires full training. We initialized our models with the
hyperparameters obtained during the random search described above. Then
we trained them until convergence, with the number of epochs varying somewhat between experiments.


\subsubsection{Results}

We carried out over 2,000 experiments as part of hyperparameter
search.  Figure~\ref{fig:hyper-search-landscapes} shows the
hyperparameter response landscapes after the first epoch for random,
GloVe, and retrofitted GloVe. The landscapes are similar for all the
embedding families, with a high mean region for lower values of the
learning rate and initial range, followed by a big plateau with low
mean and low variance for higher values of the hyperparameter pairs,
which yielded poorer performance. There is a limit boundary in both
hyperparameters delimited by constant lines around the value $3$ for
both the learning rate and the initialization range, after which the
network enters a non-learning regime.

\begin{figure}
  \includegraphics[width=\linewidth, clip=true, trim= 0 0 0 0]{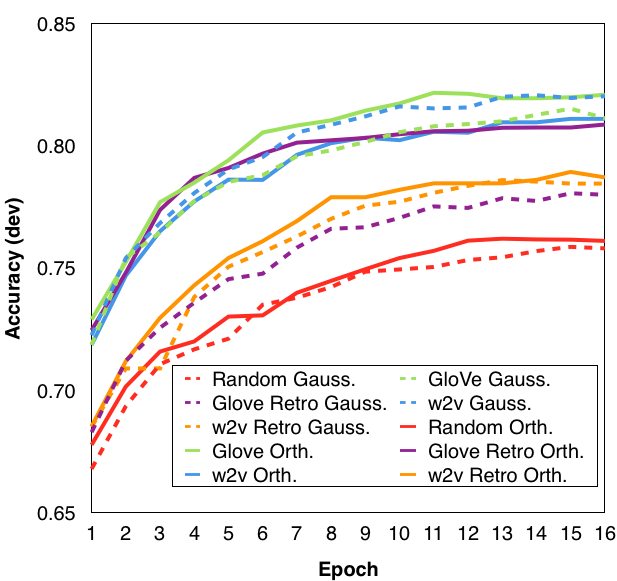}
  \caption{Accuracy on SNLI dev dataset as a function of the 
    epoch for models run until convergence.
  }
  \label{fig:snli-final-accuracy}
\end{figure}

After a single
epoch, all kinds of pretraining outperform random
vectors. Surprisingly, though, GloVe and word2vec also outperform
their retrofitted counterparts. Table~\ref{tab:snli-final} shows that
this ranking holds when the models are run to convergence, independently of whether Gaussian or orthogonal weight
initialization schemes are used. Figure~\ref{fig:snli-final-accuracy}
traces the learning curves for the different inputs and weight
intialization schemes.\footnote{Within the 160 values reported in
  Figure~\ref{fig:snli-final-accuracy}, we performed an interpolation
  on 8 of them due to an issue in our reporting system. None of these
  involved initial or final values.}
In addition, for all but random and word2vec, orthogonal
initialization is notably better than Gaussian, and
results in a mean gain of $0.8\%$ across all the pretrained
embeddings, with gains of $1.0\%$ for GloVe vectors and $2.9\%$ for
retrofitted Glove.

\begin{table}
  \centering
  \begin{tabular}[c]{r r r}
    \toprule
                   & Gaussian & Orthonormal \\
    \midrule
    random         & 75.81 & 76.11 \\
    GloVe          & 81.13 & \textbf{82.10} \\
    word2vec       & 82.04 & 81.11 \\
    retro Glove    & 78.00 & 80.88 \\
    retro word2vec & 78.46 & 78.71 \\
    \bottomrule
  \end{tabular}
  \caption{Accuracy results for the SNLI dev-set experiment.}
  \label{tab:snli-final}
\end{table}

\subsubsection{Analysis}

The SNLI experiments confirm our expectations that pretraining helps,
in that all the methods of pretraining under consideration here
outperformed random initialization. Evidently, the structure of the
pretrained embeddings interacts constructively with the weights of the
network, resulting in faster and more effective learning.
We also find that orthogonal initialization is generally better.  This
effect might trace to the distance invariance ensured by the
orthogonal random matrices. Vector orientation is important to GloVe
and word2vec, and even more important for their retrofitted
counterparts. We conjecture that keeping invariant the notion of
distance encoded by these embeddings is highly profitable.

The one unexpected result is that retrofitting hinders performance on
this task. The retrofitted information is essentially WordNet
hierarchy information, which seems congruent to the NLI task. Why does
it appear harmful in the SNLI setting? To address this question, we
conducted two simpler experiments, first with just pairs of words
(sequences of length $1$ for the premise and hypothesis) and then with
words with increasingly complex negation sequences.
These simpler experiments suggest
that the complexity of the SNLI data interacts poorly with the highly
structured starting point established by retrofitting.

\subsection{Lexical Relations in WordNet}\label{sec:lex-rel-wn}
Retrofitting vectors according to the scheme used here means infusing them with information from the
WordNet graph \cite{WordNet98,faruqui2014}. We thus expect this process to be
helpful in predicting word-level entailment relations between words.
\newcite{bowman2014} released a dataset that allows us to directly
test this hypothesis. It consists of 36,772 word-pairs derived from
WordNet with the label set
$\set{\emph{hypernym}, \emph{hyponym}, \emph{coordinate}}$. This
section reports on experiments with this dataset, showing that our
expectation is met: retrofitting is extremely beneficial in this
setting.


The model used the same settings as in the SNLI experiment, and we
followed the same procedure for hyperparameter search. We study only
random, GloVe, and retrofitted GloVe vectors to keep the presentation
simple. The dataset was split into $80\%$ training, $10\%$
development, and $10\%$ test. The runs to convergence used
$35$ epochs.

\subsubsection{Results}

As Table~\ref{tab:wordnet-final} shows, retrofitting is extremely
helpful in this setting, in terms of both learning speed and overall
accuracy. GloVe and random performed similarly, as in
\citealt{bowman2014}.

\begin{table}[tp]
  \centering
  \begin{tabular}[c]{r r}
    \toprule
    random         & 94.32  \\
    GloVe          & 94.45 \\
    word2vec       & 94.26 \\
    retro Glove    & \textbf{95.68} \\
    retro word2vec & 95.49 \\
    \bottomrule
  \end{tabular}
  \caption{Accuracy results for the WordNet experiment.}
  \label{tab:wordnet-final}
\end{table}


\subsubsection{Analysis}

These results show clearly that retrofitted vectors can be helpful,
and they also provide some clues as to why retrofitting hurts with
SNLI. The process of retrofitting implants information from external
lexical sources into the representations, modifying the magnitudes and
directions of the original embeddings (while trying to keep them close
to the original vectors). For example, the word \word{cat} is modified
so it incorporates the notion that it entails \word{animal}, by
adjusting the magnitudes and directions of these embeddings. These
adjustments, so helpful for word-level comparisons, might actually
make it harder to deal with the complexities of semantic composition.
Our next experiment explores this hypothesis.

\subsection{Lexical Relations with Negation}
To begin to explore the hypothesis that semantic composition is the
root cause of the variable performance of retrofitting, we conducted
an experiment in which we introduced different amounts of semantic
complexity in the form of negation into a word-level dataset and
carried out our usual batch of assessments.  The expectation is that, if
the hypothesis is true, the performance of retrofitted embeddings with
respect to a baseline should decrease. 

Inspired by work in natural logic \cite{maccartney2009a,Icard:2012},
we created a novel dataset based on a rich theory of negation. The
dataset begins from a set of 145 words extracted from a subgraph of
WordNet. We verified that each pair satifies one of the
relations $\set{\emph{hypernym}, \emph{hyponym}, \emph{equal}, \emph{disjoint}}$. The label set is
larger than in our previous experiments to provide sufficient logical
space for a multifaceted theory of negation. For instance, if we begin
from `p hypernym q', then negating just `p' yields `not p hypernym q',
negating just `q' yields `p neutral not q', and negating them both
reverses the original relation, yielding `not p hyponym not q'.  The
full table of relations is given in
Table~\ref{tab:negation}. Crucially, the variables `p' and `q' in this
table need not be atomic; they can themselves be negated, allowing for
automatic recursive application of negation to obtain ever larger
datasets.\footnote{The category `neutral', does not appear in the
  original dataset, but emerges as a result of the application of the
  theory of negation and quickly becomes the dominant category.}


Following the methodology of \newcite{Bowman:Potts:Manning:2015}, we
train on formulae of length $m$ and test exclusively on formulae of
length $n > m$ in order to see how well the networks generalize to
data more complex than any they saw in training. More specifically, we
train on the dataset that results from two complete negations of the
original word-level data. This yields $\approx 26k$ examples. Our
first test set was created by applying negation thrice to the original
word-level relations. Successive test sets were obtained using the
same system, up to the sixth level from the original dataset.  Because
the application of negation introduces a huge bias towards the
category `neutral', the test datasets were downsampled to
$\approx 10k$ examples with the same distribution of labels as the
training dataset.

\begin{table}
\small
\centering
\begin{tabular}{r c c c}
    \toprule
             & not-p, not-q & p, not-q & not-p, q \\
    \midrule
p disjoint q & neutral      & hyponym  & hypernym \\
p equal q    & equal        & disjoint & disjoint \\
p neutral q  & neutral      & neutral  & neutral  \\
p hyponym q  & hypernym     & disjoint & neutral  \\
p hypernym q & hyponym      & neutral  & disjoint \\
    \bottomrule
\end{tabular}
\caption{The theory of negation used to define the dataset for the negation
  experiment. `p' and `q' can either be simple words or potentially
  multiply negated, multi-word terms.}
\label{tab:negation}
\end{table}

\subsubsection{Results}

\begin{figure}[tp]
  \centering
  \includegraphics[width=0.8\linewidth]{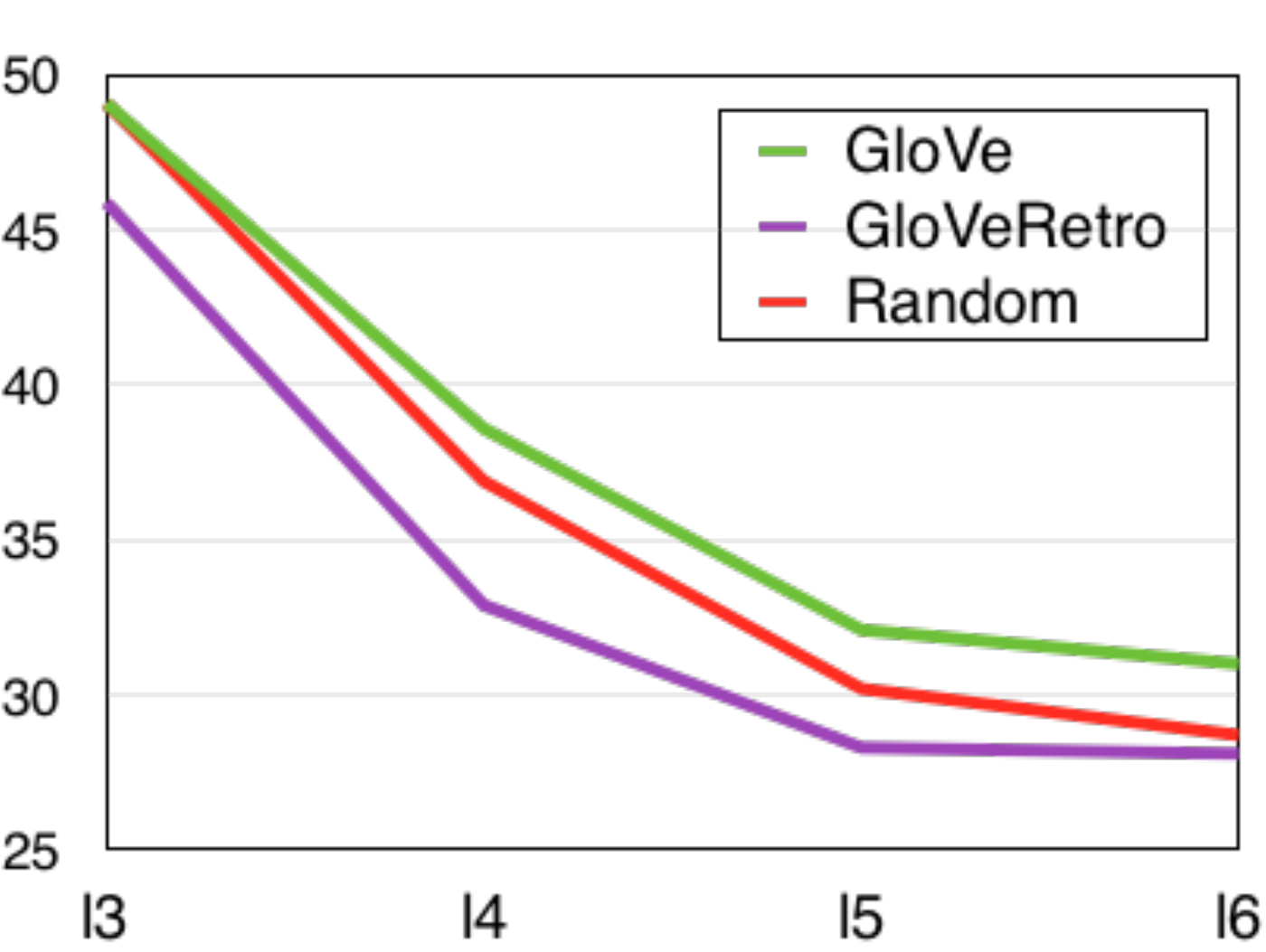}
  \caption{Accuracy on the complex negation dataset. The x-axis values
    correspond to levels of negation: For example, l3 contains terms like `not
    not not p'.}
  \label{fig:negation-l3-l6}
\end{figure}

The results are presented in Figure~\ref{fig:negation-l3-l6}. The
conclusion is very clear: the performance of retrofitted vectors drops
more when compared with the rest of the embeddings,
suggesting that compositional complexity is indeed a problem for
retrofitting. GloVe vectors again generalize best. (We expect word2vec
to be similar.)

\subsubsection{Analysis}

The drop in performance suggests that the introduction of semantic
complexity can degrade the performance of retrofitted vectors. It is
thus perhaps not surprising that the highly complex SNLI sentences
cause even deeper problems for these inputs.  More generally, these
results show that retrofitting high-dimensional embeddings is a tricky
process. As it is defined, the retrofitting algorithm seeks to keep
the retrofitted vectors close to the original vectors. However, in
high-dimensional spaces, these small changes can significantly impact
the path of optimization. It seems likely that these problems can be
mitigated by modifying the retrofitting algorithm so that it makes
more global and uniform modifications to the embedding space.

\section{Conclusion}
The central finding of this paper is that pretraining inputs with
methods like GloVe and word2vec can lead to substantial performance
gains as long as the model is configured and optimized so as to take
advantage of this initial structure. In addition, we found that, for
pretrained inputs, orthogonal random initialization is superior to a
Gaussian initialization. Random input initialization remains a common choice for some tasks.
This is perhaps justified where the training set is massive. 
%
The datasets used for neural machine translation (NMT) are often
large enough to support this approach. However, while the datasets
for other NLP tasks are growing, NMT is still unique in this
sense. The dataset used for our main evaluations (SNLI) is large by
NLP standards, and the effects of pretraining proved greater than what
can be obtained otherwise.

Our results did present one puzzle: retrofitting the vectors with
entailment-like information hindered performance for the
entailment-based NLI task. With a series of controlled
experiments, we traced this problem to the complexity of the SNLI
data. These results show that fine-tuning high-dimensional embeddings
is a delicate task, as the process can alter dimensions that are
essential for downstream tasks. Nonetheless, even
this disruptive form of pretraining led to better overall results than
random initialization, thereby providing clear evidence that
pretraining is an effective step in learning distributed
representations for NLP.

\begin{appendix}

\section{Appendix}\label{appendix:hyper}
Fixed values of less variable hyperparameters, based on the results of a coarse search:
%
\begin{itemize}\setlength{\itemsep}{0pt}
\item Gradient clipping \cite{pascanu2013}: initially set in
  $[3.0, 5.0]$, and finally to $3.0$.
  
\item Number of layers: $1$, $2$, or $4$ layers. We obtained close
  results with $2$ and $4$ layers. Although it is generally believed
  that more layers help, we fixed to $2$ to save computation time
  \cite{erhan2009difficulty,erhan2010does}.

\item Dimensionality of the embeddings: $d \in \set{50, 100, 300}$. In our experience, increasing $d$ is generally better. We chose $d=300$, as this is the largest available
  off-the-shelf distribution of GloVe and word2vec.
  
\item Retraining schedule: the epoch to start retraining the
  embeddings was originally set in $\set{1,5,8}$. Later, it
  was set to $5$ for non-random embeddings; random embeddings were
  trained since the first iteration to maximize plasticity.
  
\item Dropout \cite{zaremba2015}: set to probability $0.2$ of dropping
  the connection.
  
\item Learning rate schedule: the epoch at which the learning rate
  starts its decay was fixed to epoch $5$, with a fine-tuning rate of
  $0.8$.
  
\item Batch size: preset at $32$ and not tuned.
  
\end{itemize}

The hyperparameter search during the first epoch led us to the
optimal hyperparameters in Table~\ref{tab:snli:hyper}.  
\begin{table}[htp]
  \centering
  \setlength{\tabcolsep}{5pt}
  \begin{tabular}[c]{r cc cc }
    \toprule
     &              \multicolumn{2}{c}{Gaussian} & \multicolumn{2}{c}{Orthonormal} \\
     &               L & IR & L & IR\\

    \midrule
    random         & 1.31& 1.86 & 0.99& 0.34 \\
    GloVe          & 1.12& 1.42 & 0.85& 0.23 \\
    word2vec       & 1.16& 0.44 & 0.98& 2.06 \\
    retro Glove    & 1.57& 1.91 & 0.80& 1.35 \\
    retro word2vec & 0.64& 2.43 & 0.44& 2.45 \\
    \bottomrule
  \end{tabular}
  \caption{Optimal hyperparameters found for the SNLI experiments. 
    L: learning rate; IR: initialization range.}\label{tab:snli:hyper}
\end{table}

\end{appendix}

\section*{Acknowledgments}
We thank Lauri Karttunen and Dan Lassiter for their insights during the early phase of the research for this paper, and Quoc V. Le and Sam Bowman for their valuable comments and advice. This research was supported in part by NSF BCS-1456077 and the Stanford Data Science Initiative.

\bibliography{emnlp2016}
\bibliographystyle{emnlp2016}

\end{document}